\pdfoutput=1
\documentclass[11pt]{article}

  \usepackage[final]{other/acl}
  \usepackage{times}
  \usepackage{latexsym}
  \usepackage[T1]{fontenc}
  \usepackage[utf8]{inputenc}
  \usepackage{microtype}
  \usepackage{inconsolata}
  \usepackage{graphicx}
  \usepackage{booktabs}
  \usepackage{multirow}
  \usepackage{amsmath}
  \usepackage{amssymb}
  \usepackage{algorithm}
  \usepackage{algpseudocode}
  \usepackage{algorithmicx}
  \usepackage{listings}
  \usepackage{xcolor}

\definecolor{codegreen}{rgb}{0,0.6,0}
\definecolor{codegray}{rgb}{0.5,0.5,0.5}
\definecolor{codepurple}{rgb}{0.58,0,0.82}
\definecolor{backcolour}{rgb}{0.96,0.96,0.93}

\lstdefinestyle{mystyle}{
    backgroundcolor=\color{backcolour},   
    commentstyle=\color{codegreen},
    keywordstyle=\color{magenta},
    numberstyle=\tiny\color{codegray},
    stringstyle=\color{codepurple},
    basicstyle=\ttfamily\footnotesize,
    breakatwhitespace=false,     
    breaklines=true,                 
    captionpos=b,                    
    keepspaces=true,                 
    numbers=none,                    
    numbersep=5pt,                  
    showspaces=false,
    showstringspaces=false,
    showtabs=false,                  
    tabsize=2
}

\title{Neurocache: Efficient Vector Retrieval for Long-range Language Modeling \\ \ }

\author{Ali Safaya \hspace{16em}  Deniz Yuret \\
  \texttt{asafaya19@ku.edu.tr} \hspace{12em} \texttt{dyuret@ku.edu.tr} \\ \\
  KUIS AI Center \\
  Computer Engineering Department \\ 
  Koç University \\
}

\begin{document}
  \maketitle
  \begin{abstract}
This paper introduces Neurocache, an approach to extend the effective context size of large language models (LLMs) using an external vector cache to store its past states. Like recent vector retrieval approaches, Neurocache uses an efficient k-nearest-neighbor (kNN) algorithm to retrieve relevant past states and incorporate them into the attention process. Neurocache improves upon previous methods by (1) storing compressed states, which reduces cache size; (2) performing a single retrieval operation per token which increases inference speed; and (3) extending the retrieval window to neighboring states, which improves both language modeling and downstream task accuracy. Our experiments show the effectiveness of Neurocache both for models trained from scratch and for pre-trained models such as Llama2-7B and Mistral-7B when enhanced with the cache mechanism. We also compare Neurocache with text retrieval methods and show improvements in single-document question-answering and few-shot learning tasks. We made the source code available under: \url{https://github.com/alisafaya/neurocache}
\end{abstract}

  \section{Introduction}
\label{sec:introduction}

Recent advancements in natural language processing have been significantly driven by the development of large language models (LLMs) such as GPT-3, GPT-4, Llama, and Llama2 \cite{brown-etal-2020-gpt3, openai-2023-gpt4, touvron-etal-2023-llama, touvron-etal-2023-llama2}. While demonstrating impressive capabilities, these models are constrained by limited context window sizes. This limitation becomes apparent in tasks that require understanding long documents, such as document summarization and academic literature review, where processing hundreds of thousands of tokens is necessary.

Various methods, including sparse attention \cite{child-etal-2019-slidingwindow, beltagy-etal-2020-longformer, zaheer-etal-2020-bigbird}, have been explored to address this limitation. However, these approaches often struggle to utilize their extended contexts \cite{liu-etal-2023-lostmiddle} fully. Recent research by \citet{xu-etal-2023-retrieval-aug} shows that retrieval-augmented models with shorter contexts (4K tokens) can match the performance of models with longer contexts (16K/32K tokens), maintaining efficiency during inference. This emphasizes the potential of retrieval-augmented strategies in LLMs.

\begin{figure}[t]
    \vspace{4em}
    \includegraphics[width=\linewidth,height=\textheight,keepaspectratio]{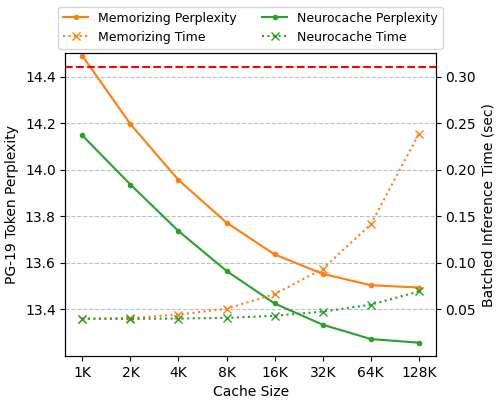}
    \caption{Performance and Scalability of Neurocache vs. Memorizing Transformers \cite{wu-etal-2022-memorizingtrans} on PG-19: The graph illustrates Neurocache's consistently lower token perplexity and faster inference times across various cache sizes on the Project Gutenberg-19 dataset, demonstrating its efficiency and scalability.}
    \label{fig:scaling}
    % \vspace{-1em}
\end{figure}

In response to these challenges, we introduce Neurocache. Neurocache employs an efficient k-nearest-neighbor (kNN) strategy for retrieving relevant past states from a compressed external vector cache. This approach is designed to optimize hidden state caching and retrieval, thereby enhancing language modeling quality and increasing inference speed.

Neurocache advances over exiting methods by reducing the cache size through the storage of compressed states, performing a single retrieval operation per token to boost inference speed, and extending the retrieval window to include neighboring states for improved language modeling and downstream accuracy. Figure~\ref{fig:scaling} illustrates the advantages of Neurocache in terms of inference speed and language modeling accuracy over methods like Memorizing Transformers \cite{wu-etal-2022-memorizingtrans}.

Our evaluation of Neurocache encompasses both models trained from scratch and established pre-trained models such as Llama2-7B and Mistral-7B \cite{touvron-etal-2023-llama2, jiang-etal-2023-mistral}, demonstrating its effectiveness in enhancing language models for downstream tasks. Specifically, we highlight Neurocache's improvements in single-document question-answering and few-shot learning tasks when compared to traditional text retrieval methods. Moreover, Neurocache's integration extends the maximum context length of these models to 128K tokens, indicating its significant impact on long-document processing.

In summary, Neurocache represents a substantial step forward in addressing the challenges of processing long documents in LLMs, offering a blend of efficiency, adaptability, and enhanced performance. Our comprehensive experiments and analysis showcase Neurocache's potential in revolutionizing the understanding of long documents in natural language processing.

  \section{Related Work}
\label{sec:related_work}

Transformers have made significant advancements in natural language processing but face challenges in processing long contexts. Various methods have been developed to extend the context window while maintaining computational efficiency \cite{huang-etal-2023-long-survey}.

Recent methods include the continued training or fine-tuning of short-context language models \cite{nijkamp-etal-2023-xgen7b, chen-etal-2023-longlora}, positional interpolation \cite{chen-etal-2023-pi}, ALiBi \cite{press-etal-2022-alibi}, and sparse and efficient attention designs \cite{child-etal-2019-slidingwindow, beltagy-etal-2020-longformer, zaheer-etal-2020-bigbird}. These approaches reflect the evolving landscape of solutions for managing extended attention windows in large language models (LLMs).

However, language models still encounter difficulties in processing longer contexts \cite{liu-etal-2023-lostmiddle}. Studies have indicated that retrieval-augmented models with shorter contexts (4K) can surpass models with longer contexts (16K/32K) in performance \cite{xu-etal-2023-retrieval-aug}.

Prominent strategies in this area are Text Retrieval and Vector Retrieval. Text Retrieval involves identifying and processing the most relevant segments of long documents. Vector Retrieval, on the other hand, integrates relevant hidden representations of the input, like hidden states or key-value pairs, into the model.

\subsection{Text Retrieval}

Text retrieval methods focus on processing relevant segments of long documents. Integrating retrieval mechanisms into language models, such as REALM \cite{guu-etal-2020-realm}, DPR \cite{karpukhin-etal-2020-dense}, RETRO \cite{borgeaud-etal-2021-retro}, and RALM \cite{ram-etal-2023-ralm}, has enhanced model performance in various tasks.

A limitation of Text Retrieval is its dependency on external retrievers for identifying relevant segments of the context, often employing algorithms like BM25 \cite{robertson-etal-2009-bm25}, Contriever \cite{izacard-etal-2022-contriever}, and others \cite{borgeaud-etal-2021-retro, ram-etal-2023-ralm, karpukhin-etal-2020-dense}.

\subsection{Vector Retrieval}

Vector retrieval methods extend the context window by incorporating relevant hidden states from an external cache of past inputs' representations.

\inlineheading{Memorizing Transformers} present a novel adaptation to the traditional transformer decoder structure for handling lengthy documents. They process documents in smaller segments and use a dynamically updated external cache to track previous key-value pairs. These models employ an approximate k-nearest-neighbor (\(k\)NN) lookup over this cache, merging dense self-attention on the current context with external-attention over retrieved key-value pairs, thus effectively extending the context length \cite{wu-etal-2022-memorizingtrans}.

\inlineheading{Unlimiformer} is a vector retrieval method, particularly suited for sequence-to-sequence models like BART \cite{lewis-etal-2020-bart}. It extends encoding length by using a \(k\)NN index over all input token hidden states, focusing on the top-\(k\) input tokens through \(k\)NN distance-based attention scores in each decoder layer's cross-attention head \cite{bertsch-etal-2023-unlimiformer}.

\subsection{Neurocache}

Neurocache is a vector retrieval method designed for processing long documents in large language models (LLMs). It employs a \(k\)NN strategy to efficiently retrieve compressed past states from an external vector cache. This approach contrasts with methods like Memorizing Transformers and Unlimiformer, particularly in terms of computational efficiency and cache size management.

Neurocache's notable features include storing compressed states to reduce cache size and performing a single retrieval operation per token, which accelerates inference speed. Additionally, it expands the retrieval window to include neighboring states, enhancing language modeling and downstream task performance.

Crucially, Neurocache shows adaptability with established pre-trained models like Llama2-7B and Mistral-7B, extending their maximum context length capabilities to 128K tokens. This adaptability demonstrates Neurocache's potential in improving long-document processing capabilities of current LLMs.

In this context, Neurocache presents a balanced approach to vector retrieval, combining efficiency and adaptability to enhance long-context processing in natural language processing models.

\begin{figure*}[!t]
    \centering
    \includegraphics[width=0.95\textwidth,height=\textheight,keepaspectratio]{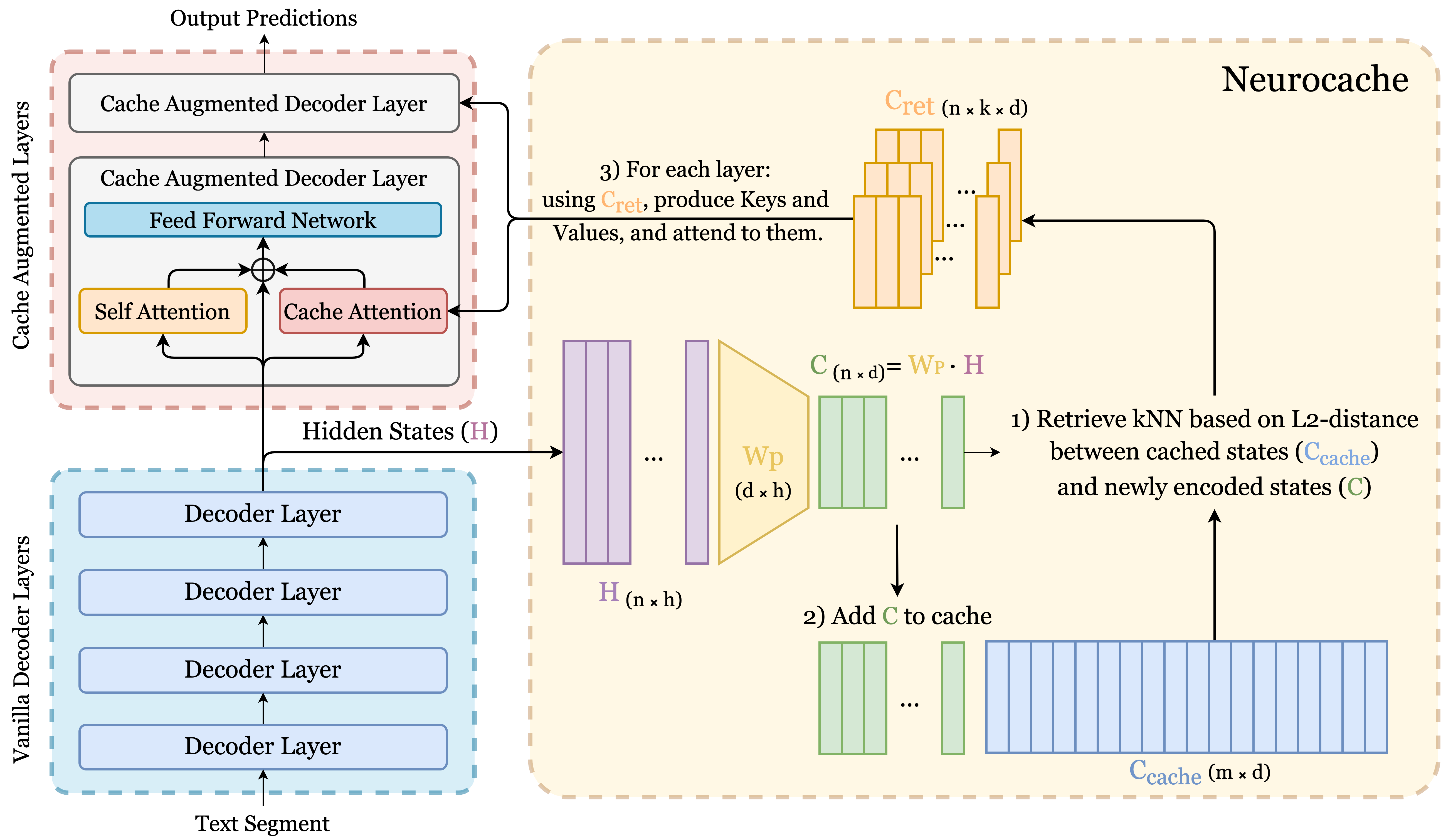}
    \caption{Documents are segmented into sequences of $n$ tokens and processed sequentially through a Transformer decoder stack. For each text segment, mid-layer hidden states $H \in \mathbb{R}^{n \times h}$ are projected into a compact representation $C \in \mathbb{R}^{n \times d}$ using a learned weight matrix $W_p \in \mathbb{R}^{d \times h}$. This projection enhances the efficiency of \(k\)NN retrieval of the most relevant past states $C_{ret} \in \mathbb{R}^{n \times k \times d}$ from the cache $C_{cache}$. These states $C_{ret}$ are used by cache-augmented layers to generate keys/values for cache attention. The output of cache attention is added to the self-attention output before being fed to the feed-forward network (FFN). Finally, the cache $C_{cache}$ is updated to include $C$ while maintaining a constant size of $m$ entries.}
    \label{fig:neurocache}
    \vspace{-1.0em}
\end{figure*}

\section{Method}
\label{sec:method}

\subsection{Neurocache Overview}

Neurocache addresses the challenge of processing long documents using Transformer decoders, leveraging a k-nearest-neighbor (\(k\)NN) search for efficient retrieval and integration of relevant past states. The process begins by segmenting the long text sequences into smaller segments, each containing \( n \) tokens, fitting the model's attention window size.

\inlineheading{State Compression:}
Text segments are sequentially processed via a Transformer decoder stack \cite{vaswani-etal-2017-attention}. At the \( r^{th} \) layer of the decoder, hidden states \( H^r \in \mathbb{R}^{n \times h} \) are acquired and subsequently projected into a compressed form \( C \in \mathbb{R}^{n \times d} \) using a learned projection matrix \( W_p \). This compression step enhances the efficiency for the subsequent \(k\)NN retrieval.

\inlineheading{State Retrieval:}
For each compressed state \(c \in \mathbb{R}^{d} \) within \(C\), we identify the top-\( k \) most similar states \( C_{ret} \in \mathbb{R}^{k \times d} \) from the cache \( C_{cache} \in \mathbb{R}^{m \times d} \). This selection is based on the L2-distance between each state in \( C \) and the states in the cache. 

\inlineheading{Cache Updating:}
The cache \( C_{cache} \) is updated with the compressed states \( C \), maintaining a fixed size of \( m \) entries. This is achieved by discarding the oldest \( n \) states, adhering to a First-In-First-Out strategy. The update occurs post-retrieval, reinforcing the commitment to retrieving only relevant past states.

% \inlineheading{State Compression and Retrieval:}
% The hidden states \( H_i^r \) of each segment \( s_i \) from the \( r^{th} \) layer are transformed into a compressed form \( C_i \). This is achieved by projecting \( H_i^r \in \mathbb{R}^{n \times h} \) into \( C_i \in \mathbb{R}^{n \times d} \) using a matrix \( W_p \). This compression significantly improves the efficiency of the \(k\)NN retrieval process that follows. In this process, we search for the top-\( k \) states \( C_{ret} \in \mathbb{R}^{n \times k \times d} \) from a cache \(C\) previous segments' states \( C_1, C_2, \ldots, C_{i-1} \) that are closest in L2-distance to \( C_i \), thereby ensuring the retrieval is confined to relevant past states and upholds sequence causality.

\inlineheading{Cache Augmented Layers:}
Using the states \( C_{ret} \) retrieved in the previous step. Starting from the \( (r+1)^{th} \) layer, the cache-augmented layers \( L^j \), where \( j > r \), integrate a specialized attention mechanism. Each layer uses unique projection matrices \( W_k^j \), \( W_v^j \), and \( W_q^j \) to generate keys \( K_{ret}^j \) and values \( V_{ret}^j \) from \( C_{ret} \), and queries \( Q^j \) from the hidden states \( H^j \). The cache attention mechanism is defined as:

\begin{equation*}
CA(Q, K_{ret}, V_{ret}) = \text{softmax} \left( \frac{Q K_{ret}^T}{\sqrt{d_{key}}} \right) V_{ret}
\end{equation*}

In this formula, \( Q \) are the queries derived from \( H^j \), while \( K_{ret}^j \) and \( V_{ret}^j \) are keys and values derived from \( C_{ret} \), with \( d_{key} \) serving as a normalization factor. The output of cache attention is processed by an output matrix \( W_o^j \) before being combined with self-attention outputs through a residual connection.

\inlineheading{Contextual Retrieval Window:}
When retrieving the top-\(k\) similar cached states, Neurocache also considers additional states surrounding these top-k states within a defined Retrieval Window. This expanded retrieval captures not only the most similar states but also their immediate neighbors, providing a richer context for the model's processing.

Consider the cached states \( C_{cache} = [c_1, c_2, \ldots, c_m] \), and a query \( q \) for which the cached states \( c_i \) and \( c_j \) are identified as the top-\( 2 \). With an even Retrieval Window size \( w \), the retrieved set would include not just \( c_i \) and \( c_j \), but also the cached states \( [c_{i-(w/2)-1}, \ldots, c_{i+w/2}] \) and \( [c_{j-(w/2)-1}, \ldots, c_{j+w/2}] \), truncated at the boundaries of the cache.

\inlineheading{Extended Cache-Attention:}
We enhance the contextual awareness of each token during the cache-attention operation by granting access to the retrievals of preceding tokens. Similar to the contextual retrieval window, this feature broadens the current token's context.

Specifically, for a token positioned at \( i \) in a sequence, denoted as \( t_i \), and with a predefined context size \( c \), the cache-attention mechanism includes not only its own retrieved states \( C_{ret}^i \) but also the states retrieved for the preceding \( c - 1 \) tokens. For example, if \( c = 4 \), the cache-attention for \( t_i \) would integrate the keys \( K_{ret}^{i-3:i} \) and values \( V_{ret}^{i-3:i} \) from tokens \( t_{i-3:i} \).

Please refer to Appendix \ref{appendix:neurocache} for more detailed description on Neurocache.

\vspace{-0.2em}
\subsection{Neurocache Adaptation}
\label{sec:method_adapt}

% Short version: Adapting pre-trained models to Neurocache involves initializing cache-attention weights from the model's self-attention projections and random initialization of \( W_p \). Low-Rank Adapters (LoRA) are integrated for efficient adaptation, and the model is fine-tuned on long document datasets.

Adapting pre-trained decoder language models for Neurocache use is a straightforward process that significantly enhances their capability to efficiently process long documents. For the layers augmented with Neurocache, denoted as \( L^j \) where \( j > r \), the adaptation involves initializing cache-attention weight matrices \( (W_k^j, W_v^j, W_q^j, W_o^j) \) by duplicating weights from the corresponding self-attention layers of the pre-trained models. Simultaneously, the projection matrix \( W_p \) is randomly initialized to transform hidden states into compact forms suitable for Neurocache retrieval.

Furthermore, we integrate Low-Rank Adapters (LoRA) \cite{hu-etal-2022-lora} into the feed-forward networks of the cache augmented layers. LoRA, introducing a minimal number of parameters, plays a key role in adapting the models to cache attention without compromising their original strengths.

During training, we freeze the original parameters of the pre-trained model and focus solely on training the newly added weights, specifically the LoRA weights, and the cache-attention weight matrices \( (W_k^j, W_v^j, W_q^j, W_o^j) \), along with the projection matrix \( W_p \). This training, using a causal language modeling objective on a corpus of long documents, enables the models to efficiently utilize the Neurocache system.

\vspace{-0.4em}
\subsection{Retrieval Overhead}

When analyzed per token, the computational overhead of retrieval in our method stems from the following components, which underline the primary computational efforts in the \(k\)NN retrieval.

\vspace{-0.4em}
\inlineheading{Distance Computation:} 
For each token, the relevance is assessed by calculating the L2-distance between the token's compressed hidden state \( c \in \mathbb{R}^{d} \) and each of the \( m \) cached states, resulting in a complexity of \( O(d \times m) \) per token, where \( d \) is the dimension of the compressed hidden state \( c \) and \( m \) is the total number of cached entries.

\inlineheading{Top-k Search Over Distances:}
Identifying the top-\( k \) closest states from these distances involves a complexity of \( O(m + k) \) for every token\footnote{We assume an algorithm with quicksort-style partitioning is used.}.

\subsection{Comparative Analysis}

The Neurocache model demonstrates computational advantage over alternatives like the Memorizing Transformer \cite{wu-etal-2022-memorizingtrans} and the Unlimiformer \cite{bertsch-etal-2023-unlimiformer} by performing only one cache query per token. This approach significantly reduces the computational burden. In contrast, the Memorizing Transformer requires multiple cache queries for each token, specifically one for every attention head. Consequently, this leads to an \( a \)-fold increase in complexity per token, both for distance computation, \( O(a \times d \times m) \), and top-\(k\) retrieval, \( O(a \times (m + k)) \), where \( a \) is the number of attention heads, and \( m \) is the cache size.

The Unlimiformer, needing \( l \times a \) queries per token, further increases retrieval complexity. For instance, a Transformer with 24 layers and 12 attention heads in the Memorizing Transformer configuration would need 12 cache accesses per token. If the Unlimiformer uses half of its layers for augmentation, as per \cite{bertsch-etal-2023-unlimiformer}, the requirement rises to \( 12 \times 12 = 144 \) cache accesses per token. Neurocache's strategy of one query per token significantly streamlines this process without compromising accuracy.

Table~\ref{table:complexity} outlines these models' retrieval frequency and cache entry size, emphasizing Neurocache's efficiency. The table compares the number of cache queries per token and each cache entry's size across the different methods.

\begin{table}[t]
\centering
\small
\begin{tabular}{lll}
\toprule[1.5pt]
Method & Retrieval & Entry \\
       & Frequency & Size \\
\midrule
Neurocache (Ours) & 1 & \( d \) \\
Memorizing Transformer & \( a \) & \( 2a \times f \) \\
Unlimiformer & \( l \times h \) & \( e \) \\
\bottomrule
\end{tabular}
\caption{Space and time complexity of methods based on cache queries per token (Retrieval Frequency) and cache entry dimensions per token (Entry Size). Here, \( d \) is the compressed dimension in Neurocache, \( a \) the number of attention heads, \( f \) head size, \( e \) hidden size, and \( l \) layers with cache attention.}
\label{table:complexity}
\vspace{-1em}
\end{table}

  \section{Language Modeling}
\label{sec:language_modeling}

% More details about the experimental setup, especially for the adaptation experiments, would be beneficial. For instance, clarifying the choice of hyperparameters and their impact on the model's performance.
% Limitations: Acknowledge any limitations of the Neurocache approach and suggest potential areas for future improvement or research.

\begin{table*}[!t]
\centering
\resizebox{0.75\textwidth}{!}{\begin{tabular}{l|c|rrrr}
\toprule[1.5pt]
\multirow{2}{*}{\textbf{Model}} & \multirow{2}{*}{\textbf{Params}} & \multicolumn{2}{c}{\textbf{PG-19}} & \multicolumn{2}{c}{\textbf{LongPile}} \\ 
                                    & & 16K & 128K & 16K & 128K \\
\midrule[1pt]
\multicolumn{6}{l}{\textit{Training from scratch}} \\
\textsc{TransformerXL}             & 184M & 14.442 & 14.442 & 15.857 & 15.857 \\
\textsc{Memorizing Transformer}    & 184M & 13.636 & 13.494 & 14.966 & 14.818 \\
\textsc{Neurocache}                & 184M & \textbf{13.511} & \textbf{13.352} & \textbf{14.425} & \textbf{14.110} \\
\midrule[1pt]
\multicolumn{6}{l}{\textit{Neurocache adaptation}} \\
\textsc{Opt-1.3B}                   & 1.3B & 12.199 & 12.199 & 19.446 & 19.446 \\
\ \ \ \ +\textsc{Neurocache}        & 1.4B & \textbf{11.306} & \textbf{11.227} & \textbf{17.626} & \textbf{17.377} \\
\midrule
\textsc{Llama2-7B}                  & 6.7B & 7.359 & 7.359 & 9.075 & 9.075 \\
\ \ \ \ +\textsc{Neurocache}        & 7.1B & \textbf{7.117} & \textbf{7.078} & \textbf{8.401} & \textbf{8.308} \\ 
\midrule
\textsc{Mistral-7B}            & 7.2B & 7.863 & 7.863 & 9.380 & 9.380 \\
\ \ \ \ +\textsc{Neurocache}        & 7.5B & \textbf{7.684} & \textbf{7.636} & \textbf{8.581} & \textbf{8.493} \\
\bottomrule[1.5pt]
\end{tabular}}
\caption{Comparison of token perplexity for different models and cache sizes on PG-19 and LongPile datasets. Neurocache outperforms Memorizing Transformer, and presents a significant reduction in perplexity across both pre-training and adaptation experiments, underscoring the adaptability to larger cache sizes.}
\label{table:perplexity}
\vspace{-1em}
\end{table*}

We assess Neurocache's effectiveness via two experimental approaches: pre-training language models from scratch and adapting established pre-trained models. For pre-training, TransformerXL  \cite{dai-etal-2019-transformer} serves as our baseline, against which we compare Neurocache and Memorizing Transformer \cite{wu-etal-2022-memorizingtrans}. In terms of adaptation, we focus on pre-trained models including OPT-1.3B, Llama2-7B, and Mistral-7B \cite{zhang-etal-2022-opt, touvron-etal-2023-llama2, jiang-etal-2023-mistral}.

% \begin{figure}[ht]
%     \centering
%     \includegraphics[width=\linewidth,height=\textheight,keepaspectratio]{figures/doc_length.png}
%     \caption{Histogram of document lengths (tokens) in LongPile and PG-19.}
%     \label{fig:doc_length}
%     \vspace{-1em}
% \end{figure}

\subsection{Datasets}

Our experiments employ two distinct raw text corpora: PG-19, a well-established benchmark for long-form language modeling, and LongPile, a diverse dataset derived from the Pile.

\inlineheading{PG-19:} This corpus comprises a collection of books written in English and published before 1919, sourced from Project Gutenberg. It is recognized as a standard benchmark for evaluating models on long-form text \cite{rae-etal-2020-compressive, wu-etal-2022-memorizingtrans, hutchins-etal-2022-blockrecurrent}.

\inlineheading{LongPile:} Extracted from the Pile corpus \cite{gao-etal-2020-pile}, LongPile features extensive documents from varied sources including "Books3," "Gutenberg (PG-19)," "OpenWebText2," "Pile-CC," and "Wikipedia (en)." The selection criterion ensures that each document surpasses 20K tokens, making it suitable for testing models' performance on longer texts.

\subsection{Pre-training}

Our baseline for pre-training is the TransformerXL model \cite{dai-etal-2019-transformer}, which we compare against Neurocache and the Memorizing Transformer \cite{wu-etal-2022-memorizingtrans}. In these experiments, both Neurocache and the Memorizing Transformer are configured with a fixed storage size of 16K during training, expanding to 128K for evaluation to assess their ability to generalize to larger storage sizes.

In Neurocache, we set the augmented layer threshold \( r \) at \( 3 * n_{layers} / 4 \), leading to the compression of outputs from the \( 9^{th} \) layer of a 12-layer model. The hidden states \( H \), originally of size \( h = 1024 \), are compressed by a factor of 4, resulting in a reduced size of \( d = 256 \). We use a retrieval window \( w = 2 \) to fetch the top-\( k \) cached states and their right neighbors for cache-attention in layers \( 10 \) to \( 12 \). Extending cache-attention to include previous tokens' retrievals with \( c = 2 \), we set \( k = 16 \), resulting in 64 neighbors in total. This setup was determined through hyperparameter optimization (details in Appendix \ref{appendix:parameters}).  Additionally, we modify the FFN dimensionality of the Neurocache from 4096, consistent with the baseline and Memorizing Transformer, to 3776 to ensure parity in model sizes.

The Memorizing Transformer, adhering to its original design \cite{wu-etal-2022-memorizingtrans}, caches key-value pairs from its \( 9^{th} \) layer. We align its retrieval setting with Neurocache by setting \( k = 64 \), thus retrieving the top-64 key-value pairs for each attention head per token.

The pre-training involves 100,000 steps with a batch size of 128 and a context size of 1,024. Adafactor \cite{shazeer-stern-2018-adafactor} is used for optimization, with a learning rate warming up over the first 1,000 steps, peaking at \( 2 \times 10^{-2} \), and then decaying to \( 1 \times 10^{-3} \). 

Neurocache's performance on the PG-19 and LongPile datasets surpasses that of the Memorizing Transformer, as evidenced by its lower token perplexities, detailed in Table~\ref{table:perplexity}. Additionally, we assess the scalability of Neurocache in comparison to Memorizing Transformers across various cache sizes. The results, illustrated in Figure~\ref{fig:scaling}, demonstrate Neurocache's computational advantage, maintaining its superior performance across different cache sizes.

\subsection{Adaptation}

We extend our adaptation strategy to pre-trained models such as OPT-1.3B, Llama2-7B, and Mistral-7B \cite{zhang-etal-2022-opt, touvron-etal-2023-llama2, jiang-etal-2023-mistral}. The adaptation process is identical to that described in Section~\ref{sec:method_adapt}, ensuring a smooth integration of Neurocache with the pre-trained model weights.

We set the rank parameter \( r \) to 16, the scale parameter \( \alpha \) to 32, and turn off bias in LoRA. Added weight matrices and adapter weights are trained on the PG-19 and LongPile datasets' training splits for 25,000 steps, employing the Adam optimizer \cite{kingma-etal-2015-adam} with a decaying learning rate of \( 1 \times 10^{-4} \). We configured Neurocache using the same settings as the pre-training experiments. This adaptation process consumes approximately 200 Nvidia A100 GPU Hours per model.

The successful adaptation is evident in the significant improvement in token perplexity on both datasets, as detailed in Table~\ref{table:perplexity}. The subsequent section discusses the impact of these improvements on zero-shot performance in downstream tasks.

  % \ \ \ \ \ \ \textsc{Position Interpolation} 
% & \textbf{26.43} & \textbf{34.70} & 40.41 & \textbf{49.01} & \textbf{21.23} & 71.00 & \textbf{43.15}  \\
% One of these models is Llama2-32K \cite{together-2023-llama-32k}, which is an extension of the original Llama2-7B which represents a long-form, full-attention model. Its context length has been increased to 32K tokens through Position Interpolation, and refined to handle longer contexts. While not as efficient as alternatives, it offers insight into the performance of full-attention models in processing extended document lengths.

\section{Downstream Evaluation}

We assess the performance of models augmented with Neurocache, particularly Llama2-7B and Mistral-7B adapted on LongPile, using seven distinct downstream tasks from the LongBench suite \cite{bai-2023-longbench}. These tasks cover a range of scenarios, including single-document question-answering (QA), multi-document QA, and few-shot learning. We utilize a zero-shot evaluation approach for the single-document and multi-document QA tasks. Conversely, in the few-shot learning tasks, a small set of examples is provided to the models, serving as part of the extended context.

\begin{table*}[!t]
\centering
\resizebox{0.75\textwidth}{!}{\begin{tabular}{l|ccc|cc|cc}
\toprule[1.5pt]
\multirow{3}{*}{\textbf{Method}} & \multicolumn{3}{c|}{\textit{Single-doc QA}} & \multicolumn{2}{c|}{\textit{Multi-doc QA}} & \multicolumn{2}{c}{\textit{Few-shot Learn.}} \\
\cmidrule{2-8}
 & \textbf{NQA} & \textbf{QSP} & \textbf{MQA} & \textbf{HQA} & \textbf{MSQ} & \textbf{TREC} & \textbf{SAMS} \\
 & F1 & F1 & F1 & F1 & F1 & Acc. & R-L\\
\midrule
\itshape Input avg. length & 35.4K & 5.3K & 8.1K & 17K & 19K & 7.8K & 11.3K \\
\midrule[1pt]
\textsc{Llama2-7B} \\
\ \ \ \ \ \ \textsc{Truncation}
& 22.19 & 28.17 & 33.39 & 33.66 & 12.30 & 67.00 & 33.00 \\
\ \ \ \ \ \ \textsc{LongLoRA}
& 21.92 & 27.58 & 30.10 & 29.17 & 11.05 & 69.50 & 30.29 \\
\ \ \ \ \ \ \textsc{Text Retrieval}
& 23.57 & 26.71 & 39.46 & \textbf{38.51} & \textbf{18.89} & 66.50 & 29.38 \\
\ \ \ \ \ \ \textsc{Neurocache} (Ours)
& \textbf{23.62} & \textbf{28.32} & \textbf{41.23} & 33.30 & 13.84 & \textbf{72.00} & \textbf{42.77} \\
\midrule
\textsc{Mistral-7B} \\
\ \ \ \ \ \ \textsc{Truncation} 
& 15.64 & 27.58 & 40.21 & 35.22 & 13.17 & 68.00 & 26.47 \\
\ \ \ \ \ \ \textsc{Text Retrieval} 
& 14.24 & 28.67 & 41.87 & \textbf{40.92} & \textbf{21.17} & 66.00 & 18.06 \\
\ \ \ \ \ \ \textsc{Neurocache} (Ours)
& \textbf{20.08} & \textbf{31.01} & \textbf{44.15} & 35.49 & 14.00 & \textbf{70.00} & \textbf{35.86} \\
\bottomrule[1.5pt]
\end{tabular}}
\caption{Zero-shot performance comparison of \textsc{Llama2-7B} and \textsc{Mistral-7B} using various long document processing methods on the \textsc{LongBench} benchmark tasks. Metrics include F1, Accuracy (Acc.), and Rouge-L (R-L). \textsc{Neurocache} excels in \textit{Single-doc QA} and \textit{Few-shot Learning} but faces challenges in \textit{Multi-doc QA} compared to text retrieval. Document lengths are provided for reference.}
\label{table:downstream}
\vspace{-1em}
\end{table*}

\subsection{Datasets}

The datasets in this evaluation present unique challenges, with average token lengths ranging from 5K to 35K, underscoring the need to process long texts effectively.

\subsubsection{Single-document QA}
\inlineheading{NarrativeQA (NQA)} is a question-answering dataset consisting of books from Project Gutenberg and movie scripts. It includes about 30 question-answer pairs per document, providing a robust test for QA systems \cite{kocisky-etal-2018-narrativeqa}.

\inlineheading{Qasper (QSP)} contains questions and answers extracted from NLP papers. This dataset offers diverse question types, such as abstractive, extractive, yes/no, and unanswerable questions, making it a comprehensive testbed for QA models \cite{dasigi-etal-2021-dataset}.

\inlineheading{MultiFieldQA (MQA)} is designed to test a model's ability to understand long contexts across various fields, including legal documents, government reports, and academic papers. It poses a challenge with its questions dispersed throughout lengthy documents \cite{bai-2023-longbench}.

\subsubsection{Multi-document QA}
\inlineheading{HotpotQA (HQA)} is a multi-document, Wikipedia-based QA dataset. It requires reading and reasoning across multiple documents and includes questions necessitating sentence-level supporting facts for complex reasoning \cite{yang-etal-2018-hotpotqa}.

\inlineheading{MuSiQue (MSQ)} focuses on multihop reasoning in QA. It constructs multi-hop questions from simpler, single-hop ones, demanding a systematic approach and detailed control over the question formation process \cite{trivedi-etal-2022-musique}.

\subsubsection{Few-shot Learning}
\inlineheading{SAMSum (SAMS)} presents a dialogue summarization challenge with its dataset of messenger-like conversations and human-annotated summaries\footnote{We use the \texttt{rouge} package: \url{https://github.com/pltrdy/rouge}}. It tests a model's ability to condense conversational data into coherent summaries \cite{gliwa-etal-2019-samsum}.

\inlineheading{TREC} serves as a dataset for few-shot learning tasks in question type classification. Models are tasked with categorizing questions into predefined categories, providing a test of their classification abilities \cite{li-roth-2002-learning}.

\subsection{Models}

In addition to Neurocache, our evaluation includes three distinct approaches for extending the input length of pre-trained language models. These approaches are Input Truncation, Text Retrieval, and Position Interpolation (PI).

\inlineheading{Truncation:}
This approach employs the original Llama2-7B and Mistral-7B models without long-context-specific modifications. Here, inputs exceeding the maximum size of 4,096 tokens are truncated from the middle following \cite{bai-2023-longbench}. This baseline serves as a reference to evaluate the effectiveness of other methods in processing extended documents.

\inlineheading{Text Retrieval:}
Contrasting with Neurocache, this approach involves selecting the most relevant text segments to include in the input, keeping the total length within the model's maximum input size. We divide the context into 200-word chunks, retrieving the top-7 chunks using Contriever \cite{izacard-etal-2022-contriever}. These chunks, along with the input, are then processed by the model. Using the top-7 chunks balances performance and the 4K token limit. This method, used in previous work \cite{bai-2023-longbench, xu-etal-2023-retrieval-aug}, differs from Neurocache, which dynamically integrates relevant information from the entire document via cache-augmented layers.

\inlineheading{Position Interpolation (PI):}
PI \cite{chen-etal-2023-pi} linearly down-scales input position indices to fit the original context window size, avoiding high attention scores that could disrupt the self-attention mechanism. LongLoRA \cite{chen-etal-2023-longlora}, leveraging PI, offers an efficient fine-tuning method to expand the context size of pre-trained models. It uses a sparse local attention mechanism, enabling computation savings while retaining performance. The fully fine-tuned LongLoRA model\footnote{\url{https://huggingface.co/Yukang/Llama-2-7b-longlora-16k-ft}}, based on Llama2-7B, extends the maximum input length to 16K tokens, aiming to assess the effectiveness of efficient full-attention methods for longer documents.

\inlineheading{Neurocache:} 
We utilize the Neurocache-adapted Llama2-7B and Mistral-7B models in our evaluation. These adaptations follow the configuration detailed in Section~\ref{sec:language_modeling} for pre-training. The models operate with a fixed cache size of 16K, accommodating the length of most datasets in our study. We split the documents into 2,048-token segments, processing them sequentially to populate the cache. Subsequently, the input, embedded within the prompt, is fed to the model, which then generates the corresponding answer.

\subsection{Evaluation Setting}

All evaluated models in this study are only pre-trained and not fine-tuned on the downstream tasks. They are assessed in a zero-shot setting, employing greedy decoding for output generation.

As outlined by LongBench \cite{bai-2023-longbench}, the model's task is to produce an answer given input and context sequences. In single-doc QA tasks, the input is a question paired with the document as context. For multi-doc QA, the input consists of multiple concatenated documents. In few-shot learning tasks, such as TREC and SAMSum, the context includes a set of examples, and the input is a question or dialogue, respectively. The input and answer are typically concise, while the context can be a long sequence extending to thousands of tokens.

If the combined length of input and context exceeds the model's maximum input capacity, only the context is truncated. This truncation is done from the middle of the context sequence, following the approach in \cite{bai-2023-longbench}. We utilize prompt templates provided by LongBench for consistency. Neurocache and LongLoRA operate with a maximum length of 16K tokens, truncating contexts longer than this limit. In contrast, the Text Retrieval method processes the entire context, regardless of length. To ensure comparability, all models are evaluated on identical hardware with a batch size of 1.

\subsection{Results}

The zero-shot evaluation results across various downstream tasks are summarized in Table~\ref{table:downstream}. We compare the performance of Llama2-7B and Mistral-7B, in their original and Neurocache-adapted forms, against other long document processing methods.

\inlineheading{Single-document QA:}
In tasks like NarrativeQA, Qasper, and MultiFieldQA, Neurocache-adapted models show superior performance, demonstrating their effectiveness in processing long contexts within single documents.

\inlineheading{Multi-document QA:}
Performance in multi-doc QA tasks, such as HotpotQA, reveals a varied picture. While Neurocache-adapted models are competitive, they fall short of Text Retrieval methods. For instance, in HotpotQA, Text Retrieval with the Mistral-7B model achieves the highest F1 score of 40.92. This finding suggests that, despite Neurocache's effectiveness in single-doc scenarios, it may be less effective in multi-doc contexts compared to text retrieval approaches.

\inlineheading{Few-shot Learning:}
In few-shot learning tasks like SAMSum and TREC, Neurocache shows strong performance, particularly indicated by improved Rouge-L scores in SAMSum. This underscores its capability to leverage few-shot examples for generating accurate summaries.

These findings illustrate the strengths and challenges of different methods in handling long documents in language models. Neurocache excels in single-document and few-shot learning scenarios, while Text Retrieval methods have an edge in multi-document tasks.

  \section{Conclusion}
\label{sec:conclusion}

This paper introduced Neurocache, an approach designed to improve long document processing in language models. Neurocache employs a k-nearest-neighbor (kNN) strategy for integrating relevant past states from compressed hidden representations, thus extending the context window of Transformer decoders. Notably, Neurocache enhances the maximum context length of models like Llama2-7B and Mistral-7B to 128K tokens.

Our findings indicate that Neurocache offers improvements in inference speed and language modeling accuracy. It demonstrates proficiency in single-document question-answering and few-shot learning, though it faces challenges in multi-document scenarios. Neurocache's competitive performance and adaptability highlight its potential utility in various applications.

In summary, Neurocache contributes to the field by enabling more efficient handling of extended contexts in existing language models. Future work may explore further optimizations for multi-document tasks and the extension of Neurocache to different model architectures and domains.

\section*{Acknowledgment}

Ali Safaya was supported by the KUIS AI Center fellowship. Deniz Yuret was partially supported by HyperBee.ai. Moreover, parts of the results reported in this paper were performed at TUBITAK ULAKBIM, High Performance and Grid Computing Center (TRUBA resources).

Ali Safaya dedicates his work to the People of Gaza.
  \section{Limitations}
\label{sec:limitations}

While Neurocache demonstrates progress in long document processing with language models, several limitations should be noted. Our evaluation is confined to datasets like PG-19 and LongPile, and tasks from the LongBench suite. These datasets, despite their diversity, might not fully represent all long-context scenarios. Performance may vary in specialized domains like technical documents or source code, which have distinct content characteristics.

A notable limitation is Neurocache's performance in multi-document scenarios, suggesting potential challenges in contexts that require integration of information from multiple sources. This aspect is crucial for applications involving comprehensive data synthesis from various documents.

In terms of bias, Neurocache depends on the underlying language models and datasets for training and evaluation. Consequently, any inherent biases in these components could influence Neurocache's outputs. An explicit analysis of model biases was not conducted in this study, highlighting an area for future exploration.

Another critical point is our reliance on a zero-shot setting for evaluation. The performance of Neurocache might differ if fine-tuning on downstream tasks or instruction datasets was employed. This limitation suggests that our current findings may not fully capture the model's adaptability and efficiency in diverse application scenarios.

In conclusion, while Neurocache presents a step forward in handling long documents in natural language processing, its effectiveness is influenced by the nature of the data, model architecture, and specific task requirements. Understanding these limitations is vital for assessing its practical applicability and guiding future improvements.
  \bibliography{bib/anthology,bib/custom}

\appendix
\appendix

\section{Optimizing Neurocache Hyperparameters}
\label{appendix:parameters}

Effective processing of long documents in Neurocache depends on the optimal tuning of various retrieval hyperparameters. To this end, we conduct a comprehensive hyperparameter search focused on language modeling performance using the Project Gutenberg-19 (PG) dataset.

Our exploration encompasses a range of values for key hyperparameters: the number of retrieved neighbors (\(k\)) with values in the set \([None, 8, 16, 32, 64, 128, 256]\), the retrieval window size (\(w\)) tested with \([1, 2, 4]\), the cache-attention context size (\(c\)) evaluated at \([1, 2, 4]\), and the encoding dimension of hidden states (\(d\)) explored across \([1024, 512, 256, 128, 64]\). This systematic investigation aims to identify the optimal configurations that enhance Neurocache's efficiency and effectiveness in handling large-scale textual data.

\inlineheading{Number of Retrieved Neighbors (\(k\)):}
The influence of \(k\), the number of retrieved neighbors, on model performance is examined. Table~\ref{table:sizeofk} shows that increasing \(k\) generally leads to a decrease in perplexity, indicating improved performance. However, the computational cost also increases proportionally with \(k\). We pick \(k = 64\) due to the diminishing returns and the increasing cost of larger values.

\begin{table}[!h]
\centering
\begin{tabular}{lcc}
\toprule[1.5pt]
\( k \) & PG (16K) & PG (64K) \\ \midrule
None & 14.739 & 14.739 \\
8   & 14.362 & 14.398 \\
16  & 14.242 & 14.259 \\
32  & 14.186 & 14.190 \\
64  & 14.117 & 14.118 \\
128 & 14.069 & 14.037 \\
256 & 14.052 & 13.988 \\
\bottomrule[1.5pt]
\end{tabular}
\caption{Perplexity for varying number of neighbors \( k \).}
\label{table:sizeofk}
\end{table}

\inlineheading{Retrieval Window Size (\(w\)):}
Adjusting \(w\) while fixing the total number of neighbors at 64, we find that a window size of \(w=2\) is optimal, as per Table~\ref{table:sizeofw}. This setting likely benefits the model's causal processing by including both the top-k entry and the subsequent one. We fix \(w=2\) for the subsequent experiments.

\begin{table}[!h]
\centering
\begin{tabular}{llcc}
\toprule[1.5pt]
\(k\) & \(w\) & PG (16K) & PG (64K) \\ \midrule
64 & 1  & 14.117 & 14.118 \\
32 & 2  & \textbf{13.720} & \textbf{13.578} \\
16 & 4  & 13.745 & 13.596 \\
% 8 & 8 & 13.781 & 13.639 \\
\bottomrule[1.5pt]
\end{tabular}
\caption{Perplexity for varying retrieval window size \( w \).}
\label{table:sizeofw}
\end{table}

\inlineheading{Attention Context Size (\(c\)):}
Table~\ref{table:sizeofc} shows that the cache-attention context size \(c=2\) achieves the lowest perplexity, indicating optimal performance when extending cache-attention to both the current and previous tokens' retrievals. We fix \(c=2\) for the subsequent experiments.

\begin{table}[!h]
\centering
\begin{tabular}{llcc}
\toprule[1.5pt]
\(k\) & \(c\) & PG (16K) & PG (64K) \\ \midrule
32 & 1  & 13.720 & 13.578 \\
16 & 2  & \textbf{13.704} & \textbf{13.564} \\
8 & 4  & 13.791 & 13.661 \\
\bottomrule[1.5pt]
\end{tabular}
\caption{Perplexity for varying context size \(c\).}
\label{table:sizeofc}
\end{table}

\inlineheading{Encoding hidden states (\(d\)):}
Finally, we assess the impact of encoding hidden states into smaller dimensions \(d\), as compared to the original \(h = 1,024\). Table~\ref{table:sizeofd} demonstrate performance degradation as smaller sizes of compression are used.

\begin{table}[!h]
\centering
\begin{tabular}{lcc}
\toprule[1.5pt]
\(d\) & PG (16K) & PG (64K) \\ \midrule
1024 & 13.704 & 13.564 \\
512  & 13.740 & 13.594 \\
256  & 13.779 & 13.641 \\
128  & 13.853 & 13.730 \\
64   & 13.983 & 13.891 \\
\bottomrule[1.5pt]
\end{tabular}
\caption{Perplexity for varying \(d\).}
\label{table:sizeofd}
\end{table}

% \inlineheading{Cache Augmented Layers:}
% Results:
% | Cache     | Attention | 16K    | 128K   |
% |-----------|-----------|--------|--------|
% | 18        | 18        | 17.875 | 17.662 |
% | 18        | 18-24     | 17.626 | 17.377 |
% | 12,18     | 12-24     | 17.343 | 17.102 |
% | 6,12,18   | 6-24      | 17.254 | 17.021 |
% | 0,6,12,18 | 0-24      | 17.240 | 17.014 |

\section{Neurocache Algorithm}
\label{appendix:neurocache}

\begin{algorithm*}
\small
\caption{Neurocache Processing}
\begin{algorithmic}[1]
\Require Long document \(D\), Segment size \(n\), Number of transformer layers \(L\), Number of lower layers \(r\), Cache memory size \(m\), Projection dimensions \(h\) to \(d\), Number of nearest states \(k\)
\Ensure Updated cache \(C_{cache}\) after processing each segment

\State Initialize cache \(C_{cache}\) with size \(m \times d\)
\State Divide document \(D\) into segments \(S = (s_1, s_2, \ldots)\)
\For{each segment \(s \in S\)}
    \State \(H^r \leftarrow\) Process \(s\) through lower \(r\) standard decoder layers
    \State \(C \leftarrow W_p \cdot H^r\) \Comment{Project hidden states to compact representation}
    \State \(C_{ret} \leftarrow\) Retrieve top-\(k\) nearest states from \(C_{cache}\) based on L2-distance to \(C\)
    \For{\(j \leftarrow r+1\) to \(L\)}
        \State \(Q^j \leftarrow W_q^j \cdot H^j\) \Comment{Generate queries for cache attention}
        \State \(K_{ret}^j \leftarrow W_k^j \cdot C_{ret}\) \Comment{Generate keys for cache attention}
        \State \(V_{ret}^j \leftarrow W_v^j \cdot C_{ret}\) \Comment{Generate values for cache attention}
        \State \(CA \leftarrow\) Apply cache attention using \(Q^j, K_{ret}^j, V_{ret}^j\)
        \State \(CA \leftarrow  CA \cdot W_o^j\) \Comment{Apply output projection for cache attention}
        \State \(H^j \leftarrow\) Combine \(CA\) with self-attention outputs and \(H^j\)
    \EndFor
    \State Update cache \(C_{cache}\) with \(C\), discard oldest if cache exceeds \(m\)
\EndFor
\end{algorithmic}
\end{algorithm*}

\begin{itemize}
\item The cache \(C_{cache}\) is initialized to store compact representations, with a maximum capacity of \(m\) entries.
\item The long document \(D\) is segmented into sequences of \(n\) tokens each.
\item Each segment \(s_i\) undergoes sequential processing through the transformer decoder layers.
\item At the middle \(r^{th}\) layer, the hidden states \(H^r\) are converted into a compact representation \(C\).
\item The nearest cached states \(C_{ret}\) to \(C\) are retrieved from \(C\) using a k-nearest-neighbor (kNN) method.
\item In each augmented layer \(j > r\), cache attention is calculated using the generated queries \(Q^j\), keys \(K_{ret}^j\), and values \(V_{ret}^j\).
\item The output of cache attention \(CA\) is merged with the self-attention outputs and subsequently processed through a feed-forward network (FFN).
\item After processing each segment, the cache \(C\) is updated with the new compact representation \(C\), and the oldest entries are discarded as needed to maintain the cache size.
\end{itemize}

The cache-attention mechanism in the augmented layers is designed to focus on the most relevant information retrieved from the cache, akin to the approach in Memorizing Transformers \cite{wu-etal-2022-memorizingtrans}. Cache-attention implementation is given in Figure~\ref{fig:cache_attention}.

\begin{figure*}[htbp]
\begin{lstlisting}[language=python]
def cache_attention(
    ret_keys, 
    ret_vals,
    queries
  ):
  # Attention computation over states retrieved from the cache.
  # ret_keys: Retrieved keys (bsz, n_queries, n_heads, n_neighbors, head_dim)
  # ret_vals: Retrieved values (bsz, n_queries, n_heads, n_neighbors, head_dim)
  # queries: Queries (bsz, n_queries, n_heads, head_dim)

  # Calculate attention weights.
  ret_attn = einsum("...qhd,...khd->...qk", queries, ret_keys)
  ret_attn = softmax(ret_attn, dim=-1)

  # Compute the weighted sum of extended values.
  attn_output = einsum("...qk,...khd->...qhd", ret_attn, ret_vals)
  return attn_output
\end{lstlisting}
\caption{This implementation showcases the cache-attention computation in the model. It calculates the attention weights through a dot product between the queries and keys, applies a softmax to obtain probabilities, and then computes the weighted sum of the extended values to generate the final attention output.}
\label{fig:cache_attention}
\end{figure*}

\end{document}